\documentclass[sigconf]{acmart}

\usepackage[show]{chato-notes}
\usepackage{balance}
\usepackage{float}

\newcommand{\gm}[1]{\textcolor{black}{#1}}
\newcommand{\er}[1]{\textcolor{black}{#1}}
\newcommand{\jl}[1]{\textcolor{black}{#1}}

\AtBeginDocument{%
  \providecommand\BibTeX{{%
    \normalfont B\kern-0.5em{\scshape i\kern-0.25em b}\kern-0.8em\TeX}}}

\setcopyright{acmlicensed}
\copyrightyear{2018}
\acmYear{2018}
\acmDOI{XXXXXXX.XXXXXXX}

\acmConference[SIGIR '24]{July 14--18,
  2024}{Washington, D.C.}
  
\acmISBN{978-1-4503-XXXX-X/18/06}

\begin{document}

\title{Zero-shot and Few-shot Generation Strategies for Artificial Clinical Records}

\author{Erlend Frayling}
\affiliation{%
\institution{University of Glasgow}
\country{United Kindgom}
}
\email{erlend.frayling@glasgow.ac.uk}

\author{Jake Lever}
\affiliation{%
\institution{University of Glasgow}
\country{United Kindgom}
  }
\email{jake.lever@glasgow.ac.uk}

\author{Graham McDonald}
\affiliation{%
\institution{University of Glasgow}
\country{United Kindgom}
}
\email{graham.mcdonald@glasgow.ac.uk}

\renewcommand{\shortauthors}{E. Frayling et al.}

\begin{abstract}
The challenge of accessing historical patient data for clinical research, while adhering to privacy regulations, is a significant obstacle in medical science. An innovative approach to circumvent this issue involves utilising synthetic medical records that mirror real patient data without compromising individual privacy. The creation of these synthetic datasets, particularly without using actual patient data to train Large Language Models (LLMs), presents a novel solution as gaining access to sensitive patient information to train models is also a challenge. This study assesses the capability of the Llama 2 LLM to create synthetic medical records that accurately reflect real patient information, employing zero-shot and few-shot prompting strategies for comparison against fine-tuned methodologies that do require sensitive patient data during training. We focus on generating synthetic narratives for the History of Present Illness section, utilising data from the MIMIC-IV dataset for comparison. In this work introduce a novel prompting technique that leverages a chain-of-thought approach, enhancing the model's ability to generate more accurate and contextually relevant medical narratives without prior fine-tuning. Our findings suggest that this chain-of-thought prompted approach allows the zero-shot model to achieve results on par with those of fine-tuned models, based on Rouge metrics evaluation.

\end{abstract}





\maketitle

\section{Introduction} \label{sec: introduction}
\gm{Clinical research is essential for improving the understanding of diseases, developing new and more effective treatments, and improving the care of patients. Access to clinical medical records, such as the hospital discharge notes and electronic health records (EHRs) ~\cite{hoerbst2010electronic, coorevits2013electronic} can aid this research to identify patterns of symptoms and drug side effects. Obtaining access to such records is challenging, due to the sensitive, personal patient information that the records contain~\cite{nurmi2019privacy}. Such challenges ultimately slow the progress of new medical discoveries that \jl{could benefit patient health}~\cite{cowie2017electronic}}. 

\gm{Developing approaches that can} alleviate privacy concerns in the clinical research space is desirable to enable easier access to EHRs such that research can be carried out more freely, leading to quicker discoveries in health-related fields. One \gm{approach that could potentially alleviate the challenges that arise from sensitive patient information} is to generate \textit{synthetic} patient records \gm{that have the same statistical distribution of terms as the real medical records but are, indeed, fake. Such synthetic medical records could then be used as a substitute for real EHRs where patient privacy barriers prevent accessing the real data~\cite{iveSynthetic}}.

Several works have explored generating synthetic EHR text using transformer-based Large Language Models (LLMs).\gm{, e.g.,}~\cite{melamudTowards,iveGeneration}. In particular, \gm{the} work by Ive \textit{et al.} \gm{\cite{iveGeneration}} showed that synthetic clinical text can be used to augment real EHR data \gm{and improve the effectiveness of LLMs} in downstream tasks~\cite{iveSynthetic}. \er{However, to prepare these models to produce synthetic EHRs, they first need to be trained on real EHR data, which brings us back to the initial issue of accessing private EHR information.}

More \gm{recently, a number of LLMs, that are pre-trained using large volumes of data and that leverage prompt inputs to discern the nature of the generative task, e.g. ~\cite{brownGpt3, touvronLlama2}, have \gm{been} shown to be effective \gm{for a} wide range of tasks. Such models do not require fine-tuning.} Being able to utilise such LLMs for generating synthetic EHR data would remove the need to gather real, hard-to-access, EHR fine-tuning data. 

In this work, we evaluate the capabilities of the Llama 2 LLM, with a variety of learning strategies, \gm{e.g. including fine-tuning, few-shot and zero-shot learning settings}, to generate synthetic clinical \gm{EHR} text. \gm{In particular, we deploy the evaluated models to generate synthetic History of Present Illness narratives \jl{from the short Chief Complaint text that summarises the main medical problem. We} compare the generated narratives to real EHRs from the MIMIC-IV dataset \cite{mimicFour}.} Furthermore, we propose a chain-of-thought (CoT) prompting strategy that can be used to guide a LLM in generating EHR content with consideration for the structure and specific content of EHRs\gm{. Our experiments show} that this CoT method can improve zero-shot and few-shot learning strategies with Llama 2 to be competitive with a fine-tuned GPT-2 model, thus reducing the need to access real EHR \gm{data, that contains sensitive patient data, when conducting clinical research}. 



\section{Related Work}\label{sec:background}
The majority of work on clinical text generation utilises the transformer-based deep learning architecture in causal language modelling tasks with auto-regressive language models~\cite{vaswaniAttention, radford2018Gpt,scholkopf2021toward}. Amin-Nedjad \textit{et al.} proposed to generate patient discharge summaries from input structured patient EHR data with GPT-2\cite{radford2019language} and showed that these can be used to train more effective Named Entity Recognition (NER) models~\cite{amin2020exploring}. Similarly, Lu \textit{et al.} showed that synthetic clinical text can be used to augment a real EHR training dataset for improved performance in re\gm{-}admission predictions tasks~\cite{lu2021textual}. Several other works \gm{have} also investigated using generated synthetic text in downstream tasks\gm{,} such as \gm{the work} of Melamud \textit{et al.}~\cite{melamudTowards} who showed that synthetic records can be used in Natural Language Inference tasks. Li \textit{et al.}~\cite{liAre} trained several autoregressive models to generate History of Present Illness subsections of EHR discharge summaries and manually annotate the \er{synthetic} records for entity mentions\gm{. Li \textit{et al.}~\cite{liAre}} showed that a more effective NER model can be trained \gm{by} using the annotated synthetic data to augment real training dataset. There also exists a significant amount of work in EHR summarisation with sequence-to-sequence models\gm{, e.g.,} \cite{RaffelT5, gaoSummarizing, palNerual, hartman2022day}\gm{. H}owever\gm{, differently from the work of \cite{RaffelT5, gaoSummarizing, palNerual, hartman2022day}, in this work} we focus exclusively on the auto-regressive task of generat\gm{ing synthetic clinical data}.

The majority of work on clinical text generation uses the Medical Information Mart for Intensive Care (MIMIC) datasets. MIMIC-III~\cite{johnsonMimic3} is a large, publicly accessible database containing detailed clinical data from patients admitted to critical care units. Recently MIMIC-IV~\cite{mimicFour} \gm{has been} released\gm{. MIMIC-IV} contains many more records \er{than MIMIC-III} \gm{and, therefore we use the MIMIC-IV dataset for our experiments. H}owever due to its recency, there \gm{has been} less \gm{prior} work utilising \gm{MIMIC-IV} for text generation tasks compared to MIMIC III. Both datasets contain a variety of structured and unstructured data\gm{,} including patient demographics, laboratory results, procedures and healthcare staff written notes. Most of the aforementioned works~\cite{amin2020exploring, lu2021textual, melamudTowards}, directly evaluate the quality of the generated clinical text using metrics that measure term overlap\gm{,} such as ROUGE score and BLEU score~\cite{linRouge, papineniBleu}, though the latter is more typically used to evaluate machine translation model performance - therefore, in this work, we use the ROUGE family of metrics to evaluate the quality of our generated synthetic records compared to gold-standard examples.

\section{Electronic Health Record Generation with LLMs}\label{sec: method}
As described in \gm{Section}~\ref{sec:background}, to generate synthetic text\gm{,} an auto-regressive language model is trained on a dataset of real text. The nature of auto-regressive models make them ideal for causal language modelling tasks where a language model \textit{models} the distribution of terms in a dataset\gm{,} such that \gm{the model} can predict what token should come next given a prior sequence \gm{of tokens} and a set of vocabulary tokens\gm{.} Equation~\ref{eqn: nexttokenpred} shows how \jl{the} probability of the next token in a sequence in calculated given an initial sequence of discrete tokens\gm{, }where \begin{math} W_{0}\end{math} is the initial context word sequence, \begin{math} W_{t}\end{math} is candidate next token,  and \begin{math} w_{1:0} = \emptyset \end{math} indicates the first word of the initial sequence.
\begin{equation}
    P(w_{1:T} \mid W_{0}) = \prod_{t=1}^{T} P(w_{t} \mid w_{1:t-1}, W_{0}) \text{ with } w_{1:0} = \emptyset
    \label{eqn: nexttokenpred}
\end{equation}
\er{In our task we model two parts of the unstructured text from EHR record in a causal language modelling task. These are:}

\begin{enumerate}
    \item Chief Complaint (CC) - a short key phrase description of an admitted patient's main medical issue\gm{.}
    \item History of Present Illness (HPI) - a longer explanation as to how the patient came into hospital for their ailment, including causes of ailments, patient's , and other notes from the hospital staff.
\end{enumerate}

\er{The objective of our task, then, is to effectively model a relationship between the CC and HPI with LLMs, so that given a CC an LLM will produce a HPI.} This way the model can be prompted to generate HPIs that may be of interest to researchers, or for use in downstream tasks as used in previous works \cite{melamudTowards, amin2020exploring,lu2021textual} . While this task could typically be achieved by fine-tuning a generative model on formatted text passages containing CCs and HPIs, we focus on developing prompting strategies to use LLMs without fine-tuning in a zero-shot and few-shot setting, to remove the need to access sensitive patient data for fine-tuning, relying instead on the parametric knowledge of the pre-trained model to generate synthetic HPIs.

\subsection{Prompting Strategies} \label{sec: inContext}
In the remainder of this section, we describe the different prompting strategies we \gm{deploy for} generating the HPI sections of EHRs from the text of a provided CC \gm{section}. We also describe the various learning strategies we \gm{deploy} prompt LLMs, including zero-shot prompting and few-shot prompting. We design prompts for the Llama 2 LLM architecture, which utilises a \textit{System Prompt} component to provide the model with additional context information about the nature of generative task of the model \cite{touvronLlama2}. We use this system prompt this propose a tailored chain of thought (CoT) prompting strategy for generating synthetic medical text.

\subsubsection{Direct Prompt Strategy}
Firstly, we propose a text prompt that includes the names of both sections of EHRs we are concerned with. The prompt is provided to the model as a single input sentence (where \textit{X} is replaced with a real Chief Complaint): 
\begin{quote}
    The Chief Complaint is: \{X\}. The History of Present Illness is:
\end{quote}
This format provides context about the kind of information being presented, namely a CC, and prompts the model to begin generating a corresponding HPI for \gm{the} provided CC. The reductive nature of this prompt formatting is grounded by the fact that clinical record data is restricted and it may not be possible to provide the model with additional information about EHRs in a closed, sensitive setting. \gm{This prompting strategy is denoted as Direct Prompt in Section~\ref{sec:results}.}

\subsubsection{Chain-of-Thought Method}

\begin{figure*}[tb]
\includegraphics[scale=0.5]{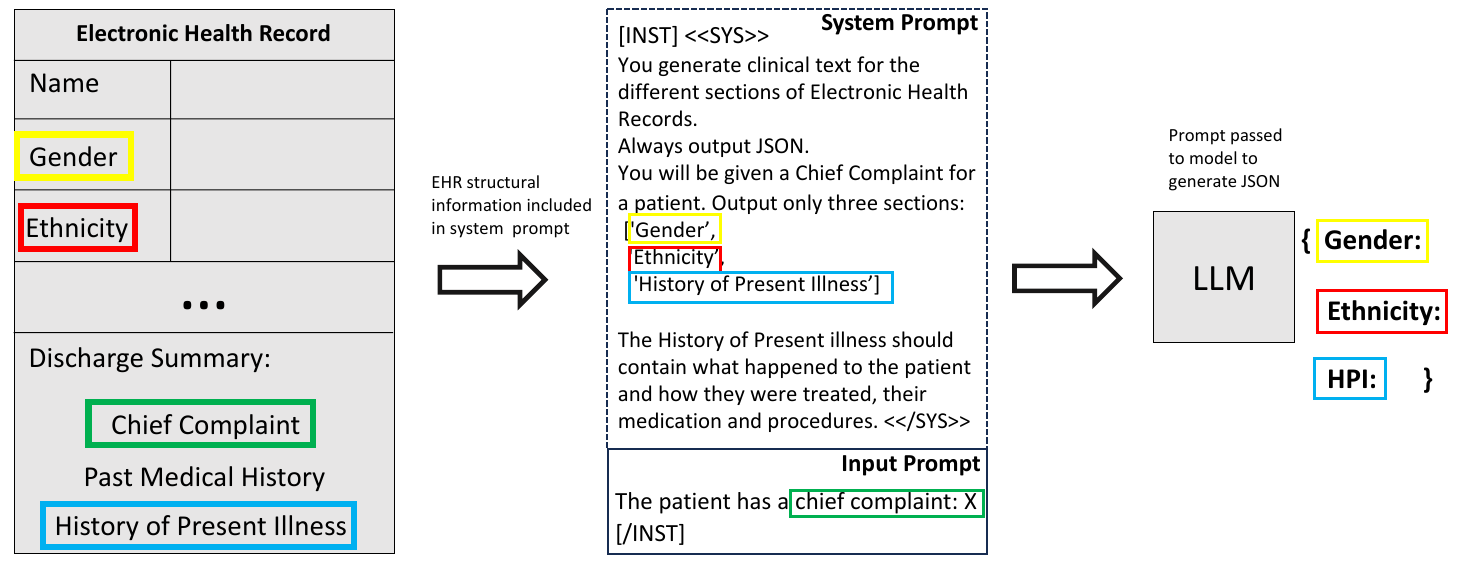}
\caption{A mock example of an EHR record (left) and its formatted data in our designed CoT Prompt (centre), separating the System Prompt from the Input Prompt. The prompt is passed to an LLM model (right) for generation.}
\label{fig:cotStruct}
\end{figure*}

Secondly, we propose a more complex prompt\gm{ing strategy based on the} Chain-of-Thought (CoT) \gm{paradigm}. \gm{CoT} can explicitly guide an LLM through multiple steps of reasoning during a task~\cite{wei2022chain}. We propose to instruct the model to generate other parts of \gm{the} EHR records\gm{,} for a given CC\gm{,} \textit{before} instructing the model to generate the HPI. Specifically, we instruct the model to first generate a gender for the patient of the provided CC, \gm{followed by} an ethnicity \gm{of the patient}, and finally the HPI. In doing so, we hypothesise that the model should use its own additional answers about these more simple concepts to generate a more realistic HPI. 

We utilise this CoT process with the System Prompt component of the Llama 2 LLM (our chosen model for prompting). The Llama 2 LLM was trained to use a \textit{system prompt} that is inserted before the users prompt. This system prompt is used to inform the model about its general task and function. \er{In this case we modify the original system prompt proposed in \cite{touvronLlama2} and instead instruct the model to generate clinical data and to output each component of the CoT instructions as JSON.} Figure \ref{fig:cotStruct} illustrates how the CoT prompt is structured and how it references the structure of an EHR, using the System Prompt before \gm{considering the} specific CC for which to generate a HPI displaying. The System prompt is presented to the model with a special \begin{math}<<SYS>> \end{math} token\gm{. This prompting strategy is denoted as [Chain-of-Thought] in Section~\ref{sec:results}.} 

\subsection{Learning Strategies} \label{sec: learnStrats}
We \gm{deploy each of} our prompt\gm{ing} strategies with three auxiliary learning strategies, i.e. how the prompt is passed to the model. Firstly, we use zero-shot prompting\gm{,} where the prompt is passed without any other \gm{contextual} information. Secondly, \gm{we additionally} pass examples of what the output is expected to be, \gm{i.e.,} few-shot learning, \gm{which} has \gm{been} shown to improve the performance of in-context learning models~\cite{brownGpt3}. We propose to use few-shot learning in two ways, first\gm{ly} \gm{by} sampling random examples to be used with a given prompt\gm{,} and secondly \gm{by providing the model with} examples that are similar to the main input prompt.

\section{Experiments} \label{sec: experiment}
In this section we describe \gm{the} experiments we perform to answer the following \gm{three} research questions:

\noindent{\textbf{RQ1}: Can a LLM achieve the same performance in generating HPIs using prompting strategies compared to fine-tun\gm{ed} LLMs?}

\noindent{\textbf{RQ2}: Does our proposed CoT prompting strategy improve \gm{the} performance of prompt-based generation with LLMs?}

\noindent{\textbf{RQ3}: How do our prompting strategies perform in \jl{zero-shot and few-shot settings?}}



    

\subsection{Experimental Setup}
\subsubsection{Dataset} \label{sec: dataset}
We used the MIMIC-IV~\cite{mimicFour} data collection to create a dataset of Chief Complaints with corresponding History of Present Illness records, we extract 7000 discharge summaries of patients that contained both a Chief Complaint \gm{(CC)} and also a History of Present Illness \gm{HPI} section in their records. Our dataset was split into a training and a test collection consisting of 6000 training samples, used to train baseline models that use fine-tuning, and 1000 test samples for evaluation of the generated HPIs. For each sample we also extract the gender and ethnicity of the patients corresponding to each CC-HPI pair.

\subsubsection{Models and Evaluation}
To answer our research questions we \gm{deploy} three transformer-based architecture models, namely GPT-2, BioGPT~\cite{luo2022biogpt} and LLaMA-2 13B (Llama). Table \ref{tab: modelInfo} provides an overview of \gm{the} learning strategies \gm{and models} we use. Notably, we use GPT-2 baseline \gm{due to} its wide use as a fine-tuning model in many different generation tasks. We choose BioGPT \gm{due to} its biomedical pre-training, which may improve performance in the clinical domain, \gm{resulting from} the similarity of clinical data to biomedical data. Finally, we use the Llama LLM  in both a fine-tuning setting and for our prompting strategies with each \gm{auxiliary} learning strategy.

\begin{table}
\caption{A comparison of chosen LLMs and the learning strategies used with each model to generating clinical texts.}
\label{tab: modelInfo}
    \centering
    \resizebox{\columnwidth}{!}{
    \begin{tabular}{c| c   c   c   c}
    \hline
    \textbf{Model} & \textbf{\# Parameters} & \textbf{Fine-Tuning} & \textbf{In-Context Strategies} \\ \hline
    GPT2 & 1.5 billion & $\checkmark$ & $\times$ \\
    BioGPT & 1.5 billion & $\checkmark$ & $\times$ & \\
    Llama 2 & 7 billion &  $\checkmark$* & $\checkmark$ \\
    \end{tabular}
    }
    \label{tab:my_label}
\end{table}

\begin{table*}[h]
  \caption{Perplexity and ROUGE Score results, evaluating generated history of present illness summary against target true history of present illness summary. Best results in each metric highlighted in bold}
  \label{tab:Rouge_table}
  \begin{tabular}{|l|c|c|c|c|c|c|} \hline  
     Strategy&Model&  Perplexity & Rouge-1& Rouge-2& RougeL&RougeL-Sum\\ \hline \hline
          &GPT2& 11.6& 0.23& 0.036& 0.116&0.233\\ \hline  
  Fine-tuned&BioGpt& 10.3& 0.264& 0.048& 0.132&0.238\\ \hline  
  &Llama2 +QLoRA& \textbf{6.16}& \textbf{0.282}& \textbf{0.052}& \textbf{0.146}&\textbf{0.246}\\ \hline \hline
  &Zero-Shot& - & 0.172& 0.028& 0.109& 0.164\\ \hline  
  Direct Prompt&Few-Shot (random)& - & 0.191& 0.025& 0.113& 0.185\\ \hline  
  &Few-Shot (similar)& - & 0.205& 0.03& 0.11& 0.20\\ \hline \hline
 & Zero-Shot& -& 0.236& 0.043& 0.126&0.220\\\hline
 Chain-of-Thought& Few-Shot (random)& -& 0.208& 0.031& 0.109&0.195\\\hline
 & Few-Shot (similar)& -& 0.228& 0.043& 0.123&0.212\\\hline 
 \end{tabular}
\end{table*}

Firstly, we fine-tune each model on our CC-HPI dataset described in Section \ref{sec: dataset}, concatenating the CC and HPI texts training samples with an added special token, <|sep|>. For the Llama model we load the model with 4-bit quantisation~\cite{dettersQLoRA} and use Low Rank Adaptation~\cite{HuLoRA} to efficiently \gm{fine-tune} the LLM for the generation task\gm{,} due to its larger size. For each model we perform 20 runs of hyperparameter tuning using Optuna~\cite{akiba2019optuna}, searching Learning Rate, Weight Decay, and Number of Epochs. We optimise for evaluation loss and use the best hyperparameter configuration to train a final model that is used in evaluation.

Secondly, we use the Llama 2 model with our prompting strategies described in Section \ref{sec: inContext}. In these cases, we use the 4-bit quantized model, without any fine-tuning. For few-shot learning, we use examples extracted from the 6000 sample strong training dataset as described in Section \ref{sec: inContext}: first randomly, and secondly using the ColBERT-PRF retriever to find similar examples~\cite{wang2023colbert}. We create a dense index of the CCs in the training dataset, and for each CC in the test dataset we retrieve the top two most relevant training dataset CCs with their associated HPIs to use as the similar examples. For our proposed direct prompt and CoT prompt we then build zero-shot, few shot (random), and few-shot (similar) datasets, with the 1000 sample test collection. For the CoT prompts we also incorporate the corresponding gender and ethnicity values for the CC and HPIs.

In total, we propose six different prompt-based generation strategies for the Llama model. These are: the direct prompt, using the zero-shot, few-shot random and few-shot similar learning strategies; and the CoT prompt, using the zero-shot, few-shot random, and few shot-similar learning strategies.

Finally, to evaluate each setup, we generate HPIs for each CC in the test collection - for each of our three fine-tuned models where we pass the CC and the special separator token <|sep|> only at inference time, and for each of our six prompt-based strategies. We compare the genreated HPIs collections with the true HPIs for each CC and calculate ROUGE scores~\cite{linRouge}. We also show perplexity score for each model after the fine-tuning process where fine-tuning is used.

\balance
\section{Results and Analysis}\label{sec:results}

Table \ref{tab:Rouge_table} shows the results of each generation strategy using fine-tuning and prompting. Outright, the QLoRa fine-tuned Llama 2 model (Llama2 + QLoRA) achieves the best performance, scoring 0.28 in Rouge-1 and also scoring best in all other ROUGE metrics. BioGPT is the next best performing model with 0.264 in Rouge-1, improving by 3.4 points over the base GPT-2 fine-tuned model. This indicates that, as expected, fine-tuned models, training on many examples of CC-HPI pairs can achieve the best performance where EHR data is available to be used for fine-tuning. The perplexity scores of the fine-tuned models reflects their scores in the ROUGE metrics, with Llama 2 achieving the lowest perplexity score.

Next, we see that using our proposed CoT method provides improvement over our direct prompt in a zero-shot setting. Comparing the two approaches we see that using CoT prompting improves the performance of generation by 6.4 points, to the extent that the zero-shot Llama 2 model's performance with CoT prompting is comparable to, and slightly better than, the fine-tuned GPT-2 model. With regards to our research questions (RQs) we now answer RQ1 and RQ2. Firstly, w.r.t RQ1, we find that using our CoT prompting strategy, a zero-shot Llama 2 13b model loaded with 4-bit quantization can outperform a GPT-2 model fine tuned on EHR data in the same generation task, however it does not achieve the performance of more sophisticated fine-tuned models like Llama2 and BioGPT. In the case of the Llama 2 fine-tuned model this is to be expected, as it is the same model architecture trained on many examples for the same task. Secondly, and w.r.t RQ2, our CoT prompting method does improve zero-shot model performance compared to using a method that does not CoT. 

\looseness -1 To answer RQ3, we analyse the results of our two prompting strategies in zero-shot and few-shot learning settings. For the direct prompt, using few-shot learning improves the performance of generation - Rouge-1 score, using few-shot learning with random examples, improves performance by 0.19, and few-shot learning similarly improves performance over the zero-shot model in Rouge-1 to 0.205, an increment of 0.033. However, for the CoT prompt, few-shot learning hinders performance when using random examples and when using similar examples. Random few-shot learning reduces performance the most, a drop in 2.8 points in Rouge-1. W.r.t RQ3 we can say that our direct prompt improves performance in our generation task. However attempting to add examples to the CoT prompting method reduces model performance.

\section{Conclusion}
\looseness -1 In this work, we evaluated the effectiveness of a Llama 2 LLM for generating representative synthetic medical records, under zero-shot, few-shot and fine-tuned settings, comparing to several state of the art fine-tuned models. Moreover, we proposed two tailored prompting strategies for generating synthetic History of Present Illness sections of Electronic Health records. Our experiments on the MIMIC-IV dataset found that the recent Llama 2 model performed best with fine-tuning. However, we also showed that our tailored Chain of Thought prompting strategy, that provides information about EHR content and which sections of EHR to generate, can boost zero-shot LLM performance to the point that is competitive with a fine-tuned GPT-2 model. We see this a step towards reducing the need to access sensitive clinical data in order to perform research in the clinical field and worthy of future research.

\section{Acknowledgements}
This work was supported by the Engineering and Physical Sciences Research Council [grant number EP/X018237/1]

\bibliographystyle{ACM-Reference-Format}
\bibliography{sample-base}

\end{document}